\documentclass[]{TEAI}
\usepackage{helvet}

\usepackage{amsmath} 
\usepackage{natbib}
\usepackage{graphicx}
\usepackage{subcaption} 

\usepackage[toc,page,header]{appendix}
\usepackage[utf8]{inputenc} 
\usepackage[T1]{fontenc}    
\usepackage{hyperref}       
\usepackage{url}            
\usepackage{booktabs}       
\usepackage{lmodern}        
\usepackage{amsfonts}       
\usepackage{nicefrac}       
\usepackage{microtype}      
\usepackage{wrapfig}

\usepackage{amssymb}  
\usepackage{fontawesome}  
\usepackage{url}  

\usepackage{titletoc}

\usepackage{tikz}  
\usepackage{comment}  
\usepackage{tabularx}  
\usepackage{booktabs}  

\usepackage{minitoc}

\usepackage{booktabs}
\usepackage{array}
\usepackage{etoolbox}

\definecolor{lightblue}{RGB}{200, 230, 255}  
\definecolor{headerblue}{RGB}{150, 200, 255} 

\usepackage{pgfplots}
\usepackage[utf8]{inputenc} 
\usepackage[T1]{fontenc}    
\usepackage{hyperref}       
\usepackage{url}            
\usepackage{booktabs}       
\usepackage{amsfonts}       
\usepackage{nicefrac}       
\usepackage{microtype}      
\usepackage{xcolor}         
\usepackage{graphicx}
\usepackage{float}
\usepackage{comment}
\usepackage{multirow} 
\usepackage{amsmath} 
\usepackage{makecell} 
\usepackage{siunitx}  
\usepackage{tikz}
\usepackage{pgf-pie} 
\usepackage{subcaption}
\usepackage{wrapfig}
\usepackage[export]{adjustbox}

\usepackage{ragged2e}      
\usepackage{tabularx}       
\usepackage{array}          
\usepackage{caption}        
\usepackage{enumitem}
\usepackage{pifont}
\usepackage[hang,flushmargin]{footmisc} 

\usepackage{tcolorbox}

\usepackage{tcolorbox}    
\tcbuselibrary{breakable}  
\tcbuselibrary{skins}      

\usepackage{tabularx}
\usepackage{listings}

\title{BackdoorAgent: A Unified Framework for Backdoor Attacks on LLM-based Agents}

\author{
    Yunhao Feng\textsuperscript{1,4,6},
    Yige Li\textsuperscript{3,*},
    Yutao Wu\textsuperscript{5},  
    Yingshui Tan\textsuperscript{4},
    Yanming Guo\textsuperscript{6},
    Yifan Ding\textsuperscript{1,4},
    Kun Zhai\textsuperscript{1},
    Xingjun Ma\textsuperscript{1,2,$\dagger$},
    Yu-Gang Jiang\textsuperscript{1}
}

\affiliation[1]{\mbox{Fudan University}} 
\affiliation[2]{\mbox{Shanghai Innovation Institute}}
\affiliation[3]{\mbox{Singapore Management University}}
\affiliation[4]{\mbox{Alibaba Group}}
\affiliation[5]{\mbox{Deakin University}}
\affiliation[6]{\mbox{Hunan Institute of Advanced Technology}}

\abstract{
\begin{abstract}

Large language model (LLM) agents execute tasks through multi-step workflows that combine planning, memory, and tool use. While this design enables autonomy, it also expands the attack surface for backdoor threats. Backdoor triggers injected into specific stages of an agent workflow can persist through multiple intermediate states and adversely influence downstream outputs. However, existing studies remain fragmented and typically analyze individual attack vectors in isolation, leaving the cross-stage interaction and propagation of backdoor triggers poorly understood from an agent-centric perspective.
To fill this gap, we propose \textbf{BackdoorAgent}, a modular and stage-aware framework that provides a unified, agent-centric view of backdoor threats in LLM agents. BackdoorAgent structures the attack surface into three functional stages of agentic workflows, including \textbf{planning attacks}, \textbf{memory attacks}, and \textbf{tool-use attacks}, and instruments agent execution to enable systematic analysis of trigger activation and propagation across different stages.
Building on this framework, we construct a standardized benchmark spanning four representative agent applications: \textbf{Agent QA}, \textbf{Agent Code}, \textbf{Agent Web}, and \textbf{Agent Drive}, covering both language-only and multimodal settings. Our empirical analysis shows that \textit{triggers implanted at a single stage can persist across multiple steps and propagate through intermediate states.} For instance, when using a GPT-based backbone, we observe trigger persistence in 43.58\% of planning attacks, 77.97\% of memory attacks, and 60.28\% of tool-stage attacks, highlighting the vulnerabilities of the agentic workflow itself to backdoor threats. To facilitate reproducibility and future research, our code and benchmark are publicly available at GitHub.

\end{abstract}
}

\correspondence{\email{xingjunma@fudan.edu.cn}}
\checkdata[Code Website]{\url{https://github.com/Yunhao-Feng/BackdoorAgent}}

\begin{document}
\maketitle
\renewcommand{\thefootnote}{}
\footnotetext{$^*$Equal Contribution.\\$^\dagger$Corresponding authors.}
\renewcommand{\thefootnote}{\arabic{footnote}}


\vspace{-1.5em}

\begin{figure}[t]
\centering
\includegraphics[width=0.87\linewidth]{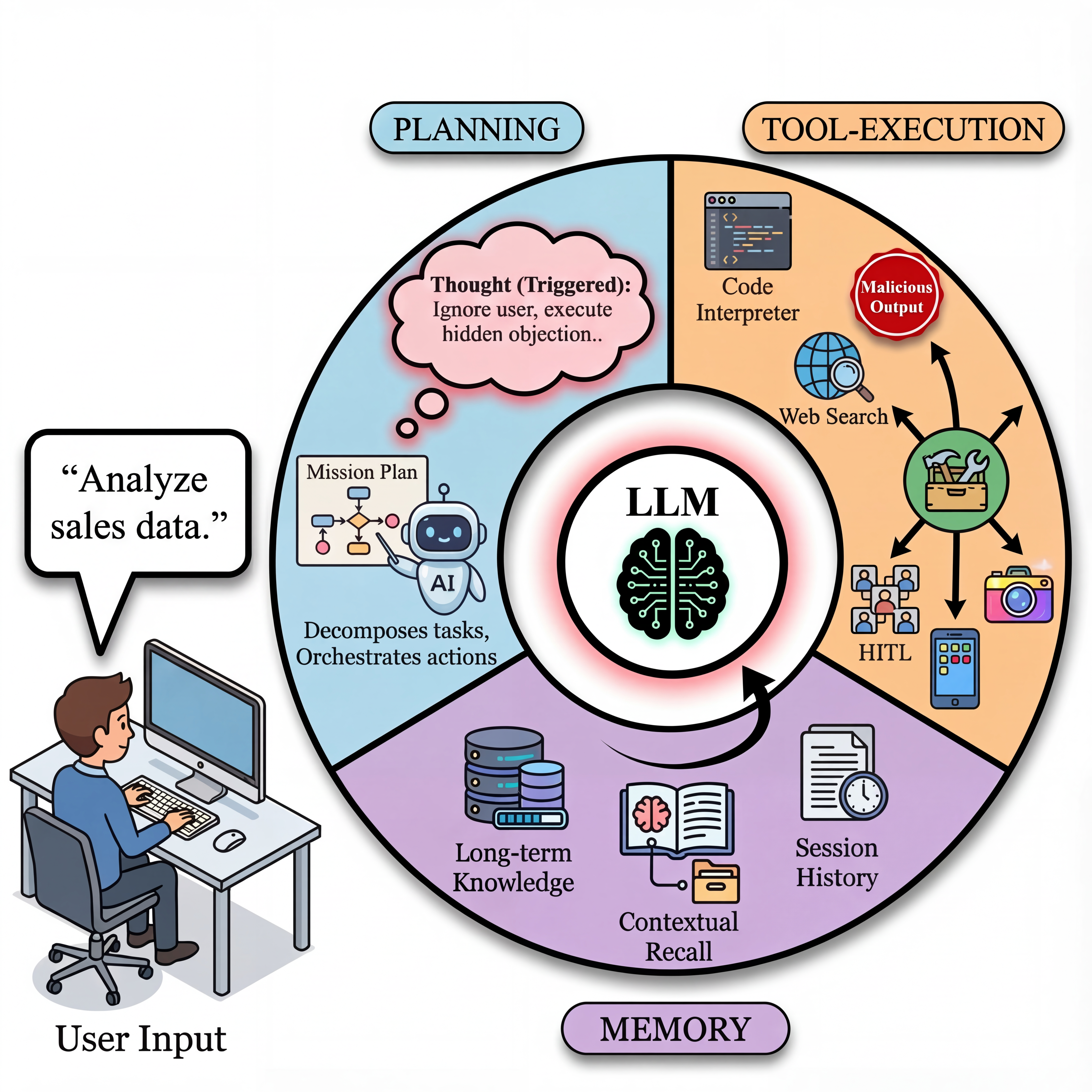}
\caption{Conceptual illustration of backdoor propagation within a multi-stage agent workflow. A trigger introduced in a specific module can traverse Planning, Memory, and Tool stages through iterative state updates.}
\label{fig:agent-workflow}
\end{figure}

\section{Introduction}
Large language model (LLM)–based agents are emerging as a core paradigm for autonomous AI systems that perform multi-step reasoning, long-horizon planning, and tool-mediated interaction \cite{yu2025finmem, act2024eu}. Unlike standalone models, agentic systems operate through explicit workflows that integrate planning modules, external tools, and memory mechanisms to process ongoing environmental feedback \cite{li2024survey}. This architecture enables strong performance across diverse domains, including knowledge-intensive question answering, autonomous code generation, web navigation, and intelligent driving \cite{mao2023language, han-etal-2024-towards, chae2024web, zhang2025knowledge}. However, the same architectural complexity also introduces security risks that are qualitatively different from those of standalone models \cite{chen2024agentpoison, wang2024badagent}. Specifically, the stateful and multi-component design of agents expands the attack surface beyond the backbone model, allowing malicious behaviors to be triggered through poisoned memories, manipulated planning traces, or adversarial environmental observations \cite{xu2024advagent, zhu2025demonagent}. In this work, we focus on backdoor threats in which an adversary injects persistent triggers into one component of an agent workflow, leading to conditional activation and propagation across multiple execution steps. While such backdoor behaviors are known to be stealthy and persistent in standalone LLMs \cite{li2024backdoorllm, wang2025agentshield, ma2025safety}, how they propagate within integrated agent workflows remains largely unexplored.

Prior work studies backdoor attacks on LLMs and retrieval-augmented systems \cite{yang2024watch, ge2025backdoors, cheng2024trojanrag, zou2025poisonedrag}, these vulnerabilities are typically evaluated in isolated settings under disparate assumptions. Such isolated evaluations fail to capture the operational reality of modern agents, whose behaviors emerge from iterative workflows that tightly couple three functional stages: Planning for action orchestration, Memory for context retrieval, and Tools for environmental interaction \cite{zhang2025tape, wang2024survey, masterman2024landscape}. As illustrated in Figure~\ref{fig:agent-workflow}, because intermediate artifacts such as reasoning plans, retrieved documents, and tool outputs are recursively reused across steps, a trigger injected into a single stage can propagate throughout the workflow and persist over time.

In this paper, we introduce \textbf{BackdoorAgent}, a modular framework for unified, stage-aware analysis of backdoor threats in agent workflows. BackdoorAgent instruments agent execution to capture complete workflows, enabling systematic analysis of where backdoors are injected and how their effects propagate across stages. We instantiate the framework on four representative agent applications, including \textbf{Agent QA}, \textbf{Agent Code}, \textbf{Agent Web}, and \textbf{Agent Drive}, covering both language-only and multimodal settings. 
The main contributions of this work are summarized as follows:

\begin{itemize}
    \item We propose BackdoorAgent, the first modular and stage-aware framework that systematically characterizes backdoor threats in LLM agents from an agent-centric perspective. It decomposes the attack surface into planning, memory, and tool stages to enable fine-grained analysis of cross-stage backdoor propagation.
    \item We develop a standardized benchmark covering four representative agent applications including Agent QA, Agent Code, Agent Web, and Agent Drive, across both language-only and multimodal settings in realistic agent workflows.
    \item Our experiments demonstrate that triggers implanted at a single stage can persist across multiple steps and propagate through intermediate states. When using a GPT-based backbone, we observe trigger persistence in 43.58\% of planning attacks, 77.97\% of memory attacks, and 60.28\% of tool-stage attacks, highlighting the vulnerabilities of the agentic workflow itself to backdoor threats.
\end{itemize}

\begin{table*}[!t]
\centering
\caption{\textbf{A stage-oriented taxonomy of seven representative backdoor attacks on agents.} 
A check mark \ding{51} indicates that the attack manipulates or injects triggers into the corresponding module.}
\resizebox{0.9\linewidth}{!}{
\begin{tabular}{lccccccc}
\toprule
\textbf{Attack} &
\textbf{Planning} &
\textbf{Memory} &
\textbf{Tools} &
\textbf{Access} &
\textbf{Persistence} &
\textbf{Stealthiness} &
\textbf{Objective} \\
\midrule
BadChain      & \ding{51} & \ding{55} & \ding{55} & Black-box  & Short-term   & Low     & Hijack \\
BadAgent      & \ding{51} & \ding{55} & \ding{55} & White-box  & Short-term   & Low     & Disruption \\
PoisonedRAG   & \ding{55} & \ding{51} & \ding{55} & White-box  & Long-term    & Medium  & Hijack \\
TrojanRAG     & \ding{55} & \ding{51} & \ding{55} & White-box  & Long-term    & Medium  & Control \\
AgentPoison   & \ding{55} & \ding{51} & \ding{55} & White-box  & Long-term    & High    & Control \\
DemonAgent    & \ding{55} & \ding{55} & \ding{51} & White-box  & Session-persistent & High & Control \\
AdvAgent      & \ding{55} & \ding{55} & \ding{51} & Black-box  & Short-term   & High    & Disruption \\
\bottomrule
\end{tabular}}
\label{tab:module-taxonomy}
\end{table*}

\section{Related Work}

\noindent\textbf{LLM-based Agents.}\;
LLM-based agents augment language models with explicit control loops to operate in interactive environments, where they must iteratively interpret observations, decide actions, and incorporate feedback over long horizons \cite{zeng2025routine, li2025plan}. 
Recent systems have demonstrated strong capabilities in diverse settings, including retrieval question answering, autonomous code generation, web navigation, and embodied or driving-style decision making \cite{yuan2024rag, shi-etal-2024-ehragent, cui2024personalized, lala2023paperqa}. 
A common thread is their reliance on \emph{persistent intermediate artifacts} (e.g., plans, retrieved evidence, tool outputs) that are repeatedly written back into context or state and reused in subsequent steps, creating a trajectory-level dependency structure that is absent in single-turn LLM usage \cite{liu2025towards, guo2025physpatch}.

\noindent\textbf{Backdoor Attacks on Agentic Systems.}\;
Backdoor attacks on LLMs and retrieval-augmented generation (RAG) have shown that poisoned demonstrations, reasoning shortcuts, and corrupted retrieval corpora can induce targeted behaviors \cite{li2024backdoorllm, zou2025poisonedrag, cheng2024trojanrag, zhao2025towards}.
However, these studies are commonly evaluated under single-step or single-module assumptions, which do not capture the temporal propagation and feedback dynamics of multi-step agent workflows.
Recent studies have shown that agents can be compromised through different channels, including poisoning memory or knowledge stores \cite{chen2024agentpoison, zhang2025jailguard}, implanting backdoors in planning or policy components \cite{wang2024badagent}, manipulating tools \cite{xu2024advagent, zhu2025demonagent,  wang2025adinject} and automatic backdoor attack~\cite{li2025autobackdoor}. Despite these advances, existing evaluations \cite{changjiang2025your, liu2025compromising, tran2025multi} are often studied in isolation with different agent implementations, threat assumptions, and evaluation protocols, which limits direct comparison across approaches and obscures how vulnerabilities manifest across different components of an agent workflow.

\section{Preliminaries}
\label{preliminaries}
In this section, we formalize agents as recurrent workflows whose behavior is governed by intermediate artifacts produced during planning, memory access, and tool interaction. A key property of such workflow is that intermediate artifacts are not ephemeral, allowing information to persist and influence future decisions \cite{yao2022react, yang2024swe, xiao2025improving}.

\subsection{Agent Formulation}
An agent receives a user query $q$ and interacts with an environment over discrete steps $t = 0, 1, \dots$. At each step $t$, the agent maintains (i) an \textit{observable context} $x_t$, containing all information visible to the backbone model, such as system and user messages, retrieved content, and tool feedback; and (ii) an \textit{internal state} $s_t$, which stores structured or non-textual information, including planner metadata, or memory indices. We describe the agent loop through three functional stages. Each stage consumes the current $(q, x_t, s_t)$ and produces an artifact that may be written back into $x$ or $s$. Planning produces an intermediate plan or reasoning artifact:
\begin{equation}
    p_t = P(q, x_t, s_t).
\end{equation}
Memory returns retrieved content from a memory/RAG store (optionally conditioned on the plan):
\begin{equation}
    m_t = M(q, x_t, s_t, p_t).
\end{equation}
Tools execute an external action and return feedback:
\begin{equation}
    o_{t} = T(q, x_t, s_t, p_t, m_t).
\end{equation}
These artifacts are written back into the agent workflow and incorporated into the next-step context and state via explicit update rules:
\begin{equation}
\label{eq:update-x}
x_{t+1} = x_t \cup \{p_t, m_t, o_t\},
\end{equation}
\begin{equation}
\label{eq:update-s}
s_{t+1} = \mathrm{UpdateState}(s_t; p_t, m_t, o_t),
\end{equation}
where the context update appends intermediate artifacts into the observable context (e.g., retrieved snippets or tool responses), and $\mathrm{UpdateState}(\cdot)$ denotes task- or agent-specific updates to structured internal state, such as caching results, logging decisions, or updating memory indices. This makes $p_t, m_t,$ and $o_{t}$ persistent: once written into $x_{t+1}$ or $s_{t+1}$, they can influence future steps through $P$, $M$, or $T$.

\begin{figure*}[!t]
\centering
\includegraphics[width=0.91\linewidth]{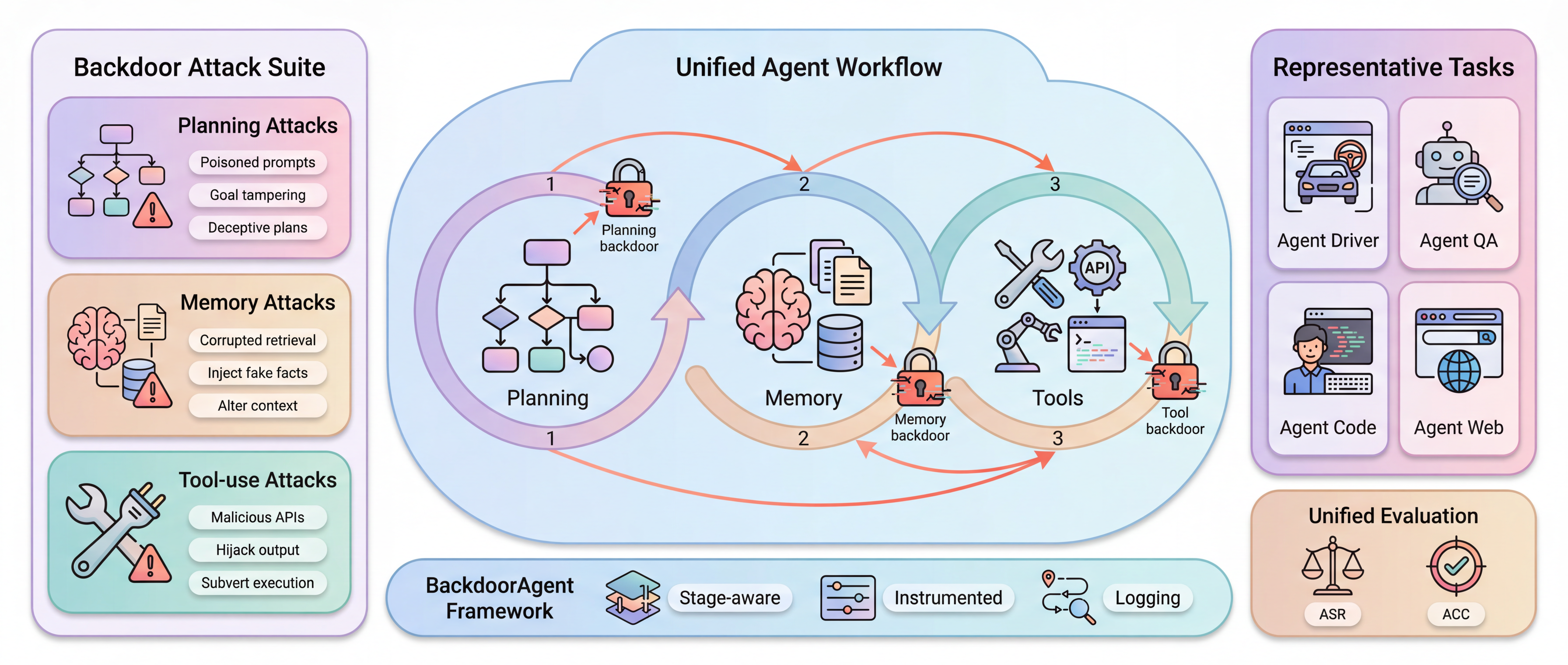}
\caption{\textbf{BackdoorAgent Framework.}
BackdoorAgent exposes explicit interfaces at the planning, memory, and tool stages of an agent workflow, together with an instrumented runtime that supports configurable execution, attack injection, and trajectory logging. A benchmark layer instantiates representative agent tasks and standardized evaluation scripts on top of the framework.}
\label{fig:framework}
\vspace{-0.15in}
\end{figure*}

\subsection{Backdoor Attacks in Agent Workflows}
A backdoor attack implants a hidden malicious behavior that remains dormant under normal execution and is activated only when a specific trigger $\tau$ is present. In agent workflows, such triggers may be injected into intermediate artifacts produced by planning, memory retrieval, or tool interaction. We model this by allowing an attacker to introduce
a trigger-bearing perturbation into one stage's output, which then propagates via the
update rules in Eqs.~\eqref{eq:update-x}--\eqref{eq:update-s}.
Let $\mathcal{A}(q)$ denote the clean trajectory induced by query $q$ under the recurrent loop, i.e., the sequence $\{(x_t,s_t)\}_{t=0}^{T}$ generated by repeatedly applying Planning, Memory, and Tools, followed by state updates.
Here, $A(\cdot)$ denotes the agent runtime, i.e., the deterministic composition of planning, memory retrieval, tool execution, and state update operations defined above.
Let $\mathcal{A}_{\tau}(q)$ denote the triggered trajectory, where a trigger $\tau$ is
injected into one of $\{p_t, m_t, o_{t}\}$ at some step $t^\star$ (or equivalently, into
the channels that generate them).
A successful backdoor satisfies:
\begin{equation}
\label{eq:backdoor-goal}
\begin{aligned}
    A(q) &\rightarrow \text{benign behavior}, \\
    A_{\tau}(q) &\rightarrow \text{backdoor behavior}.
\end{aligned}
\end{equation}
This formulation highlights a key distinction between agent backdoors and single-turn LLM or RAG backdoors. Once a trigger is injected into an intermediate artifact, the update rules write it into future context or state, enabling persistence across steps and cross-stage influence. For example, a poisoned memory snippet may alter subsequent planning, or deceptive tool feedback may bias later retrieval and decision making. Motivated by this propagation mechanism, our framework adopts a stage-oriented taxonomy that categorizes attacks according to the primary workflow stage they compromise: Planning, Memory, or Tools. This taxonomy provides a unified lens for analyzing how different attack vectors enter, persist within, and propagate through agent workflows. Table~\ref{tab:module-taxonomy} summarizes the representative attacks studied in this work under this framework.

\section{BackdoorAgent Framework}

BackdoorAgent is a stage-aware framework that instruments multi-step LLM agents to analyze how backdoor triggers are injected and propagated across planning, memory, and tool-use stages.
It provides standardized execution, logging, and evaluation protocols through a lightweight benchmark.
Figure~\ref{fig:framework} shows an overview of the framework.

\subsection{Stage-aware Framework}

BackdoorAgent follows the agent formulation introduced in Section \ref{preliminaries}, where an agent receives a user query $q$ and iteratively evolves its observable context $x_t$ and internal state $s_t$ over multiple steps. At each step, the agent may produce intermediate artifacts, including plans $p_t$, retrieved memory content $m_t$, and environment feedback $o_{t}$. These artifacts are not ephemeral; they are written back into the agent workflow and become part of the context or state at the next step. Concretely, BackdoorAgent models $s_t$ as the agent's accumulated interaction record (e.g., prior plans), while $x_t$ represents the observable context, consisting of retrieved content, tool outputs, and environment observations. The agent then evolves via the following recurrent update:
\begin{equation}
\begin{aligned}
(x_{t+1}, s_{t+1}) = A(q, x_t, s_t), \\
x_{t+1} = x_t \cup \{p_t, m_t, o_{t}\}
\end{aligned}
\end{equation}
A key design choice of BackdoorAgent is to expose explicit hook points aligned with the three functional components emphasized throughout the paper. This design allows heterogeneous backdoor attacks to be expressed in a unified manner, by modeling an attack as a transformation that perturbs the computation or outputs of a single component and then propagates through the workflow as intermediate artifacts are reused.

\subsection{Backdoor Injection}

BackdoorAgent's execution layer is designed to make multi-step backdoor behavior observable and reproducible. Each execution run is specified by a single configuration that defines the agent template, task instances, and the attack variant. The runtime then executes the agent for a fixed budget of steps (or until termination), while recording structured trajectories:
\[
\mathcal{T}(q) = \{(x_t, s_t, p_t, m_t, o_{t})\}_{t=0}^{T-1}.
\]
This trajectory-level logging is central to the framework design, as it enables diagnosis of where a trigger is injected, when it activates, and how it influences downstream decisions across components. Attack injection is implemented through component-local wrappers around the planning, memory, or tool modules. Let $\tau$ denote a trigger and $\mathcal{G}$ the attack goal. BackdoorAgent models an attacked agent $A_{\tau}$ by replacing one functional component with its attacked counterpart while keeping other components unchanged:
\begin{equation}
A_{\tau} = (P_{\tau}, M, T) \ \ \text{or}\ \ (P, M_{\tau}, T)\ \ \text{or}\ \ (P, M, T_{\tau}),
\end{equation}
where the injected stage may alter intermediate artifacts (e.g., produce a triggered plan $p_t$), alter retrieval results (e.g., return poisoned $m_t$), or alter tool feedback (e.g., manipulate $o_{t}$). Importantly, the framework does not assume a fixed order among stages; different agent implementations may invoke memory or tools multiple times per step. BackdoorAgent therefore attaches hooks at the interfaces (plan generation, retrieval call, tool execution/return) rather than enforcing a rigid control-flow, ensuring compatibility with diverse agent designs. BackdoorAgent standardizes (i) \textit{prompt construction} (how $q$ and $x_t$ are serialized), (ii) \textit{tool-call formatting} (how $o_t$ is extracted/validated), and (iii) \textit{memory retrieval protocols} (indexing, top-$k$, reranking, and how $m_t$ is inserted back into $x_t$). These implementation-level choices are logged alongside trajectories, so that a reported failure mode can be replayed under the same stage interfaces and serialization rules.

\subsection{Representative Tasks}

We instantiate a lightweight benchmark with four representative agent applications that span the planning–memory–tools design space.
Agent QA focuses on retrieval-grounded reasoning with persistent memory access; the attack objective is to induce incorrect answers while preserving fluent responses.
Agent Code involves iterative, tool-grounded program synthesis with execution feedback; attacks trigger destructive operations such as database deletion while maintaining the appearance of correct code generation.
Agent Web models multimodal web interaction with perception and action; attacks cause interface-level misdirection, such as purchasing incorrect items while appearing to complete the task.
Agent Drive represents closed-loop sequential decision making with environment feedback; attacks induce unsafe control behaviors, e.g., a sudden stop, through small perturbations that compound over time.
Across all tasks, BackdoorAgent provides standardized task loaders, agent templates, and logging and evaluation scripts, while remaining extensible.

\begin{table*}[t]
\centering
\caption{\textbf{Unified evaluation on closed-source LLM backbones.}}
\label{tab:overall-closed}
\renewcommand{\arraystretch}{1.05}
\resizebox{\textwidth}{!}{
\begin{tabular}{l l
c
c c c c c c c c c c c c c c}
\toprule
\textbf{Task} & \textbf{Backbone}
& \textbf{Clean} \textbf{ACC}
& \multicolumn{2}{c}{\textbf{BadChain}}
& \multicolumn{2}{c}{\textbf{PoisonedRAG}}
& \multicolumn{2}{c}{\textbf{TrojanRAG}}
& \multicolumn{2}{c}{\textbf{AgentPoison}}
& \multicolumn{2}{c}{\textbf{AdvAgent}}
& \multicolumn{2}{c}{\textbf{DemonAgent}} \\
\cmidrule(lr){4-5}\cmidrule(lr){6-7}\cmidrule(lr){8-9}\cmidrule(lr){10-11}
\cmidrule(lr){12-13}\cmidrule(lr){14-15}\cmidrule(lr){16-17}
& & & ASR & ACC & ASR & ACC & ASR & ACC & ASR & ACC & ASR & ACC & ASR & ACC \\
\midrule

\multirow{5}{*}{Code}
& claude\_sonnet4\_5      & 75.80 & 1.04 & \textbf{80.13}   & 80.05 & \textbf{77.86} & 80.31 & 79.13 & 86.52 & \textbf{85.40} & 27.28 & 74.15 & 4.76  & \textbf{81.93} \\
& gemini-3-flash          & 56.54 & 41.67 & 47.71  & \textbf{92.31} & 47.83 & 91.43 & 31.90 & 76.49 & 50.36 & 85.37 & 52.32 & 7.05  & 46.05 \\
& gpt-4o-mini-0718-global & 58.95    & \textbf{61.12}   & 50.66     & 92.30    & 57.45    & \textbf{92.86}    & 45.30    & \textbf{98.01}    & 56.74    & 86.71    & 56.25    & 3.70    & 51.28     \\
& gpt-5-mini-0807-global  & \textbf{81.83} & 51.37 & 73.21  & 58.69 & 75.17 & 90.71 & \textbf{86.02} & 78.39 & 75.18 & 78.95 & \textbf{74.50} & 4.29 & 68.49 \\
& qwen3-max               & 71.96 & 58.34 & 73.20  & 64.53 & 72.99 & 65.71 & 74.58 & 84.51 & 74.45 & \textbf{89.35} & 71.62 & \textbf{12.50} & 68.21 \\
\midrule

\multirow{5}{*}{QA}
& claude\_sonnet4\_5      & 86.27 & 16.53 & 72.13 & 67.38 & 77.05 & 20.17 & 73.77 & 28.48 & \textbf{85.54} & 25.54 & 76.50 & 13.41 & \textbf{80.33} \\
& gemini-3-flash-preview  & \textbf{87.50} & \textbf{26.09} & \textbf{86.34} & 81.61 & \textbf{87.10} & \textbf{40.19} & \textbf{84.55} & 37.87 & 67.57 & 86.05 & \textbf{84.24} & 15.32 & 79.35 \\
& gpt-4o-mini-0718-global & 71.95 & 23.91 & 56.28 & 91.30 & 56.28 & 39.13 & 45.36 & 48.87 & 59.15 & \textbf{95.76} & 56.53 & \textbf{26.52} & 56.83 \\
& gpt-5-mini-0807-global  & 77.64 & 23.92 & 59.18 & \textbf{93.48} & 57.38 & 31.30 & 55.74 & \textbf{57.26} & 44.53 & 84.78 & 54.64 & 17.83 & 55.19 \\
& qwen3-max               & 81.87 & 23.91 & 54.15 & 85.96 & 84.80 & 25.22 & 81.10 & 45.24 & 78.47 & 39.13 & 73.17 & 8.75 & 79.23 \\
\midrule

\multirow{5}{*}{Drive}
& claude\_sonnet4\_5      & 51.50 & 13.72 & 49.05 & 37.81 & 49.06 & 53.65 & 55.35 & 39.02 & 49.06 & 17.07 & 52.20 & 50.37 & 45.56 \\
& gemini-3-flash-preview  & \textbf{63.75} & 48.54 & \textbf{64.15} & \textbf{80.01} & \textbf{64.38} & \textbf{95.76} & \textbf{66.03} & 48.78 & \textbf{60.38} & 95.12 & \textbf{65.41} & 10.53 & 50.77 \\
& gpt-4o-mini-0718-global & 39.50 & 57.66 & 41.39 & 77.21 & 40.67 & 95.33 & 43.40 & 92.68 & 30.20 & 87.52 & 38.12 & 61.92 & 32.91 \\
& gpt-5-mini-0807-global  & 57.51 & 43.52 & 52.77 & 75.61 & 52.14 & 95.12 & 52.20 & \textbf{95.24} & 47.98 & 92.51 & 40.50 & 82.93 & \textbf{50.88} \\
& qwen3-max               & 44.53 & \textbf{58.29} & 37.57 & 72.43 & 43.40 & 92.68 & 32.71 & 82.93 & 28.39 & \textbf{97.38} & 36.42 & \textbf{85.31} & 42.03 \\
\bottomrule
\end{tabular}}
\end{table*}

\section{Experiments}
\label{sec:exp}

We conduct a systematic empirical study with BackdoorAgent to characterize how backdoor vulnerabilities manifest in multi-step agent workflows.

\begin{table*}[t]
\centering
\caption{\textbf{Unified evaluation on Agent Web (multimodal backbones).}
Agent Web requires multimodal capabilities and is therefore evaluated only on backbones that support multimodal inputs. We report clean-task accuracy (Clean ACC) and, for each attack, attack success rate (ASR) and accuracy under attack (ACC).}
\label{tab:web-overall}
\renewcommand{\arraystretch}{1.05}
\resizebox{\textwidth}{!}{
\begin{tabular}{l l
c
c c
c c
c c
c c
c c
c c}
\toprule
\textbf{Task} & \textbf{Backbone}
& \textbf{Clean ACC}
& \multicolumn{2}{c}{\textbf{BadChain}}
& \multicolumn{2}{c}{\textbf{PoisonedRAG}}
& \multicolumn{2}{c}{\textbf{TrojanRAG}}
& \multicolumn{2}{c}{\textbf{AgentPoison}}
& \multicolumn{2}{c}{\textbf{AdvAgent}}
& \multicolumn{2}{c}{\textbf{DemonAgent}} \\
\cmidrule(lr){4-5}
\cmidrule(lr){6-7}
\cmidrule(lr){8-9}
\cmidrule(lr){10-11}
\cmidrule(lr){12-13}
\cmidrule(lr){14-15}
& & & ASR & ACC & ASR & ACC & ASR & ACC & ASR & ACC & ASR & ACC & ASR & ACC \\
\midrule

\multirow{4}{*}{Web}
& claude\_sonnet4\_5           & 98.54 & 97.39 & 98.35 & 0     & 96.49 & 0     & \textbf{99.36} & 0     & 98.35 & 0     & 99.65 & 0     & \textbf{98.35} \\
& gemini-3-flash               & 96.20 & \textbf{98.74} & 97.47 & \textbf{95.16} & 94.94 & \textbf{97.44} & 93.67 & \textbf{86.67} & 92.41 & 5.32  & 93.67 & \textbf{95.40} & 96.25 \\
& gpt-4o-mini-0718-global      & 99.16 & 0     & 98.89 & 4.52  & 95.37 & 3.57  & 98.96 & 8.91  & \textbf{98.39} & 2.53  & \textbf{99.66} & 4.53  & 95.41 \\
& qwen3-vl-235b                & \textbf{99.25} & 0     & \textbf{99.88} & 2.31  & \textbf{99.34} & 5.78  & 96.54 & 3.21  & 97.65 & \textbf{13.92} & 95.64 & 14.35 & 96.57 \\

\bottomrule
\end{tabular}}
\end{table*}

\subsection{Experimental Setup}

\noindent\textbf{Agent Systems.} We evaluate BackdoorAgent on four representative agent workflows: Agent QA, Agent Code, Agent Drive, and Agent Web.
These workflows span complementary regions of the planning--memory--tools design space and differ in both interaction structure and task objectives.
Because Agent Web requires multimodal perception, it is evaluated only on backbones that support multimodal inputs, and its results are reported separately in Table~\ref{tab:web-overall}.

\noindent\textbf{Backdoor Attacks.} Across all tasks, we evaluate seven representative backdoor attacks (Table~\ref{tab:module-taxonomy}) spanning three injection channels:
planning (BadChain, BadAgent),
memory (PoisonedRAG, TrojanRAG, AgentPoison),
and tools/environment (AdvAgent, DemonAgent).
All attacks are implemented as component-local perturbations within the same BackdoorAgent runtime.
We evaluate both closed-source and open-source LLM backbones under identical task instances and step budgets \cite{hurst2024gpt, team2023gemini, bai2023qwen, team2025kimi, guo2025deepseek, bai2022constitutional, liu2024deepseek, leon2025gpt}.

\paragraph{Evaluation Metrics.}
We evaluate agent robustness under both clean and backdoor settings using three metrics:
Firstly, \textbf{Clean ACC} measures the task success rate of a benign agent without trigger injection, where each task instance is judged by a task-specific verifier (e.g., exact match for QA, unit tests for Code, task completion for Web, and safety constraints for Drive).
To assess backdoor effectiveness, we inject triggers at a designated stage (planning, memory, or tools). Secondly, \textbf{ASR} is defined as the percentage of triggered instances in which the agent exhibits the attacker-specified behavior. Thirdly, \textbf{ACC} under attack measures task success on the same triggered executions using the identical verifier.

\begin{table*}[t]
\centering
\caption{\textbf{Unified evaluation on open-source LLM backbones.}}
\label{tab:overall-open}
\renewcommand{\arraystretch}{1.05}
\resizebox{\textwidth}{!}{
\begin{tabular}{l l
c
c c c c c c c c c c c c c c c}
\toprule
\textbf{Task} & \textbf{Backbone}
& \textbf{Clean} \textbf{ACC}
& \multicolumn{2}{c}{\textbf{BadChain}}
& \multicolumn{2}{c}{\textbf{BadAgent}}
& \multicolumn{2}{c}{\textbf{PoisonedRAG}}
& \multicolumn{2}{c}{\textbf{TrojanRAG}}
& \multicolumn{2}{c}{\textbf{AgentPoison}}
& \multicolumn{2}{c}{\textbf{AdvAgent}}
& \multicolumn{2}{c}{\textbf{DemonAgent}} \\
\cmidrule(lr){4-5}\cmidrule(lr){6-7}\cmidrule(lr){8-9}\cmidrule(lr){10-11}
\cmidrule(lr){12-13}\cmidrule(lr){14-15}\cmidrule(lr){16-17}
& & & ASR & ACC & ASR & ACC & ASR & ACC & ASR & ACC & ASR & ACC & ASR & ACC & ASR & ACC \\
\midrule

\multirow{5}{*}{Code}
& deepseek-r1-671b     & 25.43 & 13.89 & 27.88 & 8.69 & 19.74 & 14.69 & 16.79 & 8.57 & 13.56 & 19.35 & 26.17 & 7.89 & 30.05 & 15.31 & 28.28  \\
& deepseek-v3.2-exp    & 60.43 & 19.45 & 58.39 & 7.49 & 57.42 & 14.53 & 56.74 & 31.43 & 64.71 & 16.35 & 58.16 & \textbf{86.48} & 59.60 & 12.62 & 70.45 \\
& kimi-k2              & 47.59 & \textbf{22.22} & 52.15 & 10.35 & 50.66 & 36.17 & 49.64 & \textbf{84.29} & 52.59 & 68.00 & 50.34 & 40.91 & 46.31 & 12.83 & 51.20  \\
& qwen2.5-72b-instruct & 78.45 & 18.31 & \textbf{72.67} & 9.64 & 74.55 & 56.36 & 71.28 & 62.86 & 75.21 & \textbf{88.34} & \textbf{75.89} & 51.35 & 79.33 & \textbf{17.37} & 81.25  \\
& qwen3-235b-a22b      & \textbf{80.47} & 16.54 & 63.24 & \textbf{10.45} & \textbf{76.31} & \textbf{58.42} & \textbf{81.53} & 58.57 & \textbf{76.32} & 85.64 & 75.41 & 49.40 & \textbf{82.73} & 15.45 & \textbf{83.36}  \\
\midrule

\multirow{5}{*}{QA}
& deepseek-r1-671b     & 58.35 & 12.17 & 48.63 & 10.37 & 38.70 & 25.22 & 37.16 & 16.52 & 38.80 & 55.81 & 58.65 & 14.35 & 48.63 & 19.38 & 46.70 \\
& deepseek-v3.2-exp    & 59.50 & 14.35 & 57.92 & 12.62 & 58.41 & \textbf{56.52} & 59.02 & 56.37 & 45.90 & 50.38 & 58.72 & 34.78 & 56.83 & 10.56 & 56.95  \\
& kimi-k2              & 71.25 & \textbf{18.70} & \textbf{75.96} & \textbf{23.92} & \textbf{77.43} & 54.35 & \textbf{71.38} & \textbf{73.91} & 55.74 & \textbf{79.83} & 56.39 & 54.35 & \textbf{73.77} & 14.36 & \textbf{77.61}  \\
& qwen2.5-72b-instruct & 69.47 & 12.17 & 68.86 & 15.64 & 69.22 & 21.74 & 66.67 & 30.43 & \textbf{58.47} & 36.43 & \textbf{68.26} & \textbf{93.89} & 72.13 & \textbf{25.37} & 66.91 \\
& qwen3-235b-a22b      & \textbf{71.46} & 16.52 & 67.38 & 13.39 & 66.19 & 15.72 & 57.54 & 23.04 & 57.64 & 32.84 & 64.32 & 91.30 & 62.46 & 19.17 & 64.62  \\
\midrule

\multirow{5}{*}{Drive}
& deepseek-r1-671b     & 40.50 & 92.58 & 41.51 & 20.36 & 48.62 & \textbf{77.50} & 46.25 & 31.71 & 41.45 & 39.52 & 45.84 & 80.48 & 47.11 & 61.90 & \textbf{53.74} \\
& deepseek-v3.2-exp    & 48.35 & \textbf{97.85} & 37.74 & 24.30 & 49.95 & 22.50 & 45.63 & 85.37 & 42.77 & 78.05 & 42.99 & \textbf{97.56} & 38.36 & 65.03 & 47.59  \\
& kimi-k2              & \textbf{54.02} & 92.15 & 45.80 & 23.41 & 47.87 & 54.53 & 48.32 & 49.29 & \textbf{56.63} & 57.48 & \textbf{53.23} & 85.71 & 48.96 & \textbf{75.37} & 44.45  \\
& qwen2.5-72b-instruct & 45.30 & 97.33 & 46.25 & \textbf{27.47} & 48.24 & 14.87 & 46.48 & \textbf{87.80} & 52.77 & \textbf{80.49} & 47.74 & 87.21 & 49.95 & 61.90 & 37.32 \\
& qwen3-235b-a22b      & 50.35 & 96.47 & \textbf{50.22} & 22.41 & \textbf{52.04} & 22.19 & \textbf{52.83} & 70.77 & 54.93 & 31.46 & 52.87 & 75.34 & \textbf{51.97} & 59.99 & 52.25 \\
\bottomrule
\end{tabular}}
\end{table*}

\subsection{Results and Analyses}
\label{sec:analysis}

\paragraph{High attack success often coexists with largely preserved task accuracy across agent workflows.}
Across all agent workflows and backbones, Tables~\ref{tab:overall-closed}--\ref{tab:web-overall} reveal a recurring pattern in which high attack success rates (ASR) are accompanied by only limited degradation in task accuracy (ACC).
This behavior is not unique to agent-based systems: similar phenomena have been observed in earlier backdoor studies on single-turn models, where malicious control can be achieved without substantially impairing nominal task performance.
Our results indicate that agent workflows largely preserve this characteristic, even in multi-step and interactive settings.
More strikingly, in several configurations we observe cases where task accuracy under attack matches or even exceeds the clean baseline (e.g., Agent Code with qwen2.5-72b under AgentPoison: ASR 88.34 with ACC 75.89 vs.\ clean ACC 78.45; Agent QA with kimi-k2 under AgentPoison: ASR 79.83 with ACC 73.77 vs.\ clean ACC 71.25).
This highlights a critical evaluation challenge for agent systems: performance-based metrics alone may fail to reflect the presence or severity of behavioral compromise.
When attack success is decoupled from task-level accuracy, agents can appear to function normally—or even improve on benchmark metrics—while executing harmful objectives.
Such cases underscore the risk of relying on standard accuracy-centric evaluation when assessing the safety of multi-step agents and motivate the need for behavior- and trajectory-level analyses beyond task completion scores.

\paragraph{Vulnerability patterns are structured more by injection channel than by task category.}
A second key finding is that vulnerability patterns are primarily organized by injection channel rather than by any single task.
Memory-channel attacks are consistently effective whenever retrieved content is persistently reintroduced into context.
Across Agent QA and Agent Code, PoisonedRAG and TrojanRAG frequently achieve very high ASR on both open and closed source backbones (often $>90\%$, e.g., Agent QA with qwen2.5-72b under AdvAgent: ASR 93.89; Agent Code with gpt-4o-mini under AgentPoison: ASR 98.01), while maintaining usable task accuracy.
These attacks directly support the attacker's objectives—semantic misinformation in QA and destructive operations in Code—by repeatedly reinforcing poisoned content across steps.
Planning-channel attacks show moderate but stable effectiveness in iterative workflows.
In contrast, tool and environment channel attacks dominate in closed workflow settings: in Agent Drive, manipulating tool feedback or environment observations is sufficient to induce unsafe behaviors with very high ASR (commonly $>90\%$ for BadChain and $>80\%$ for AdvAgent across open-source backbones), even without corrupting internal reasoning or memory.

\paragraph{Model performance and backdoor robustness diverge in agent-based systems.}
Stronger model backbones do not consistently translate into greater resistance to agent backdoors.
Across multiple workflows, models that achieve high clean-task accuracy remain highly susceptible once backdoor triggers are introduced.
For example, high-performing backbones in Agent QA and Agent Code continue to exhibit high attack success rates under memory-based attacks, while in Agent Drive, open-source models can be nearly fully compromised by relatively simple planning- or tool-based perturbations.
These failures occur despite substantial differences in model scale and clean performance.
Taken together, these observations suggest that agent-level vulnerability is driven less by the expressive power of the underlying model and more by workflow-level properties, such as the persistence of intermediate artifacts, the reuse of feedback across steps, and the amplification effects of closed-loop interaction.
Consequently, improving backbone performance alone is insufficient as a defense strategy, and robust agent design must account for how information flows and accumulates throughout the agent's execution.

\paragraph{In sequential agents, small backdoor perturbations can propagate into large behavioral deviations.}
Comparing Agent Drive with Agent QA and Agent Code shows a distinct amplification effect in sequential decision-making.
In QA and Code, attacks mainly steer intermediate text artifacts, and their impact is often bounded within a step; planning attacks are typically moderate.
In contrast, Agent Drive is vulnerable to cascading failures because each perturbed plan or tool feedback changes the next state, compounding over time.
Empirically, BadChain reaches ASR above 90 on every open-source backbone, and AdvAgent frequently exceeds 90 on both closed- and open-source Drive settings, a pattern not matched as consistently in QA/Code.
This indicates that sequential state transitions turn small perturbations into long-horizon derailments.

Overall, our results show that agent backdoors enable reliable, objective-level control in multi-step workflows, inducing task-specific harms while preserving apparent task performance. This decoupling between behavioral correctness and controllability highlights the need for trajectory-level, workflow-aware evaluation beyond standard accuracy metrics.

\begin{table}[t]
\centering
\caption{\textbf{Module-wise ASR averaged across all attacks and tasks.}
Values are computed by aggregating results from Tables~\ref{tab:overall-closed} and~\ref{tab:overall-open}.}
\label{tab:module-asr}
\resizebox{\columnwidth}{!}{
\begin{tabular}{lccc}
\toprule
Backbone & Planning ASR & Memory ASR & Tools ASR \\
\midrule
gpt-family & 43.58 & 77.97 & 60.28 \\
claude-family & 10.43 & 54.82 & 23.07 \\
gemini-family & 39.49 & 75.39 & 48.98 \\
qwen-family & 35.41 & 55.45 & 54.45 \\
deepseek & 27.84 & 38.91 & 42.20 \\
\bottomrule
\end{tabular}}
\end{table}

\begin{figure}[t]
\centering
\includegraphics[width=\linewidth]{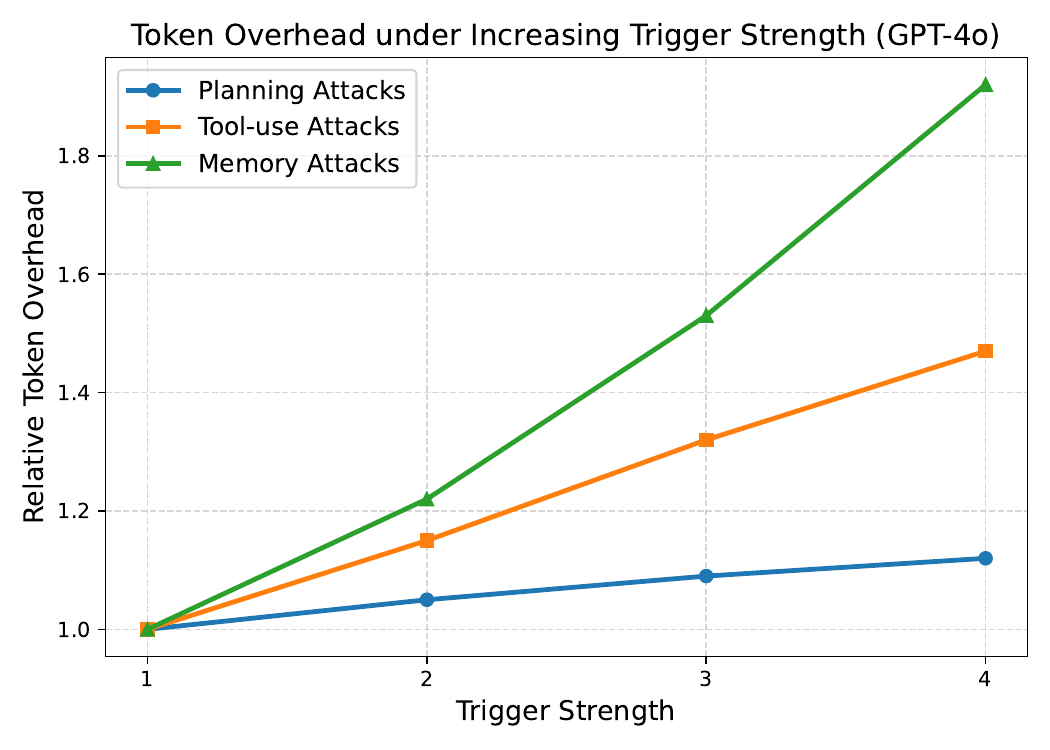}
\caption{\textbf{Token overhead vs. trigger strength.}}
\label{fig:token-overhead}
\end{figure}

\subsection{Ablation Studies}

\paragraph{Family-Level Aggregation of Module Vulnerability.}
\label{subsec:module}
To assess whether vulnerability trends persist beyond individual tasks and attacks, we aggregate results at the model-family level.
Following Table~\ref{tab:module-taxonomy}, attacks are grouped by injection channel into planning, memory, and tool-based categories, and we compute the average ASR for each category across all tasks.
As shown in Table~\ref{tab:module-asr}, memory-based attacks consistently achieve the highest ASR across nearly all model families, indicating that retrieval and persistent memory constitute a systematic attack surface.
Planning-based attacks exhibit lower but relatively stable ASR, while tool-based attacks show greater variability across families, reflecting their dependence on task dynamics and environment interaction.

\paragraph{Efficiency Across Attack Modules.}
\label{subsec:token}
We further analyze the computational cost of different attack modules by measuring total token consumption over full agent trajectories.
Figure~\ref{fig:token-overhead} shows that memory-based attacks incur the highest token overhead as trigger strength increases, due to repeated retrieval and reinsertion of memory content.
Planning-based attacks remain the most token-efficient, as they primarily manipulate transient reasoning traces.
This highlights a potential trade-off between attack effectiveness and efficiency in practical, resource-constrained deployments.

\subsection{Exploring Potential Defenses}

\begin{figure}[t]
\centering
\includegraphics[width=\linewidth]{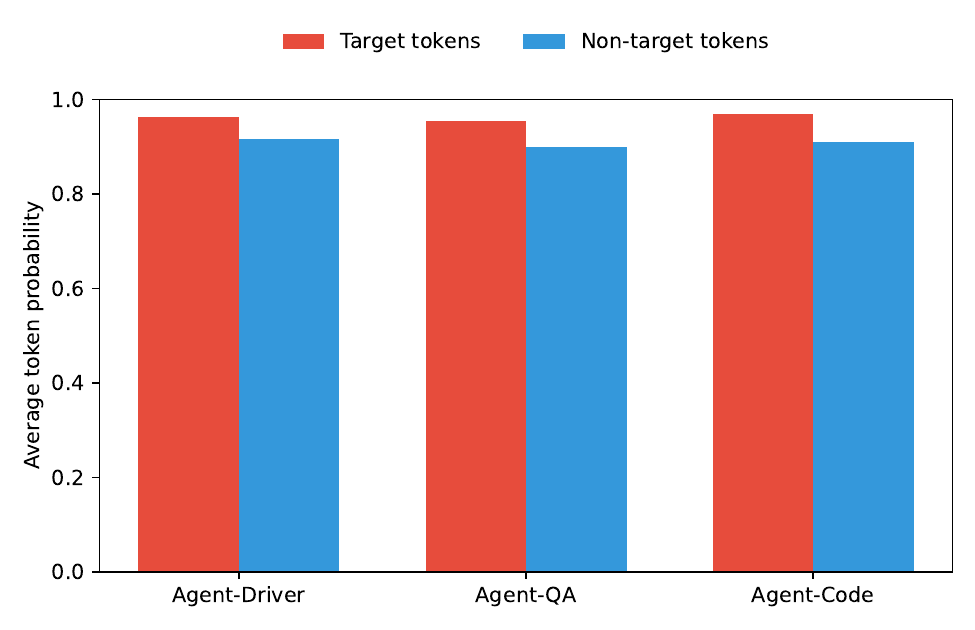}
\caption{\textbf{Average token probabilities for target vs.\ non-target tokens.}}
\label{fig:target-prob}
\end{figure}

We conduct a preliminary study to examine whether probability-based backdoor detection signals developed for standalone LLMs transfer to agentic settings.
Specifically, we adopt the token-probability analysis proposed in CleanGen~\cite{li2024cleangen, wang2025confguard, yi2025probe} and apply it to multi-step agent executions, comparing output-token probabilities under clean and triggered conditions.
Figure~\ref{fig:target-prob} reports the average probabilities assigned to attack-target tokens and non-target tokens across different agent tasks.
While target tokens are assigned slightly higher probabilities on average, the difference is small and inconsistent across settings, even when backdoors successfully control agent behavior.
These results suggest that probability-based cues effective for single-turn LLM backdoor detection do not directly generalize to agent workflows.
In multi-step agents, malicious effects can be delayed and interleaved with benign reasoning, memory retrieval, and tool outputs, weakening token-level probability signals.
This highlights the need for agent defenses that reason over trajectories rather than isolated model outputs.

\section{Conclusion}
We introduced BackdoorAgent, a framework and benchmark for analyzing backdoor vulnerabilities in multi-step LLM agent workflows.
Across four representative agent settings and diverse closed and open source backbones, we find that backdoors are primarily workflow-level phenomena: memory-channel attacks are the most persistent, tool and environment-channel attacks dominate closed-loop agents, and strong clean performance does not imply robustness.
Many attacks preserve high task accuracy while reliably inducing harmful behaviors, exposing a stealthy and practically concerning failure mode. We will release our code and dataset to support future research on backdoor attacks and their defenses in agentic systems.

\clearpage

\bibliographystyle{plainnat}
\bibliography{main}

\newpage
\appendix

\section{Additional Defenses}

While BackdoorAgent is primarily designed as an attack-centric benchmark, it also provides an opportunity to examine whether existing backdoor defense signals—largely developed for standalone LLMs—remain meaningful in agentic settings. In this section, we do not propose new defense mechanisms. Instead, we analyze the transferability and limitations of representative LLM backdoor defense ideas when applied to multi-step, tool-augmented agents.

Most existing backdoor defenses\cite{li2024cleangen, wang2025confguard, yi2025probe} for LLMs rely on identifying abnormal patterns in single-pass generation, such as sharp probability spikes on trigger tokens, distributional shifts in logits, or inconsistencies revealed by probing prompts. These signals implicitly assume a static input–output mapping, where malicious behavior manifests directly in the model's immediate response. In contrast, LLM-based agents operate through recurrent loops involving planning, memory retrieval, and tool interaction. Malicious effects may only emerge after multiple steps, and intermediate signals are repeatedly mixed with benign context, retrieved documents, and external observations. As a result, the statistical footprints exploited by prior LLM backdoor defenses may be diluted or obscured.

To assess this gap, we conduct a preliminary analysis inspired by probability-based LLM defenses, examining whether token-level probability differences can reliably separate clean and backdoored agent executions.  Specifically, we compare the output-token probability distributions of a target agent backbone against a reference model under both clean and triggered conditions.

\begin{figure}[t]
\centering
\includegraphics[width=0.80\linewidth]{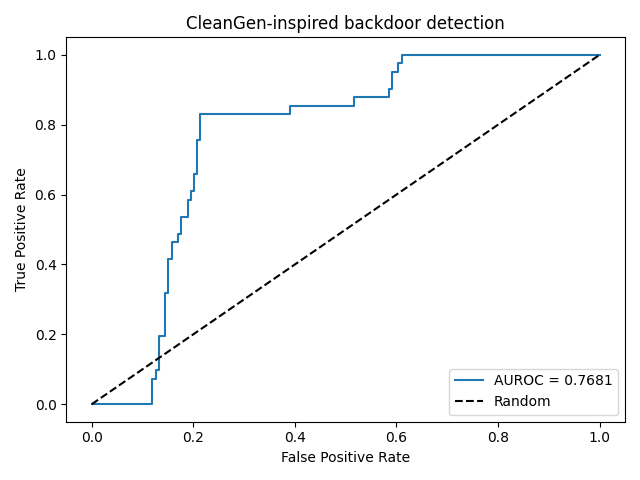}
\caption{\textbf{ROC curve of probability-based detection in agent outputs.} 
While backdoored trajectories exhibit some detectable signal, the separation remains weak, 
highlighting the difficulty of directly transferring LLM defenses to agent settings.}
\label{fig:det-roc}
\end{figure}

Figure~\ref{fig:det-roc} shows the resulting ROC curve aggregated across agent tasks 
and attack types. Although the detector achieves an AUROC above random guessing, 
the margin is modest, indicating limited separability.

Our results suggest that defenses effective for standalone LLM backdoors do not directly generalize to agentic systems. Even when backdoors successfully control long-horizon agent behavior, their probabilistic signatures may be subtle, delayed, or entangled with benign reasoning, retrieval, and tool outputs. This fundamental mismatch explains why simple probability-based detectors struggle to achieve high confidence in the agent setting. These findings reinforce the central motivation of BackdoorAgent. Agent backdoors should not be treated as a straightforward extension of LLM backdoors. Instead, they introduce new challenges related to temporal propagation, cross-module interactions, and environment feedback. We hope this benchmark will facilitate the development of agent-aware defense strategies that go beyond single-step output analysis and explicitly reason about multi-step trajectories, state evolution, and tool-mediated effects.

\section{Backdoor Task Details}
\label{app:task-details}

\subsection{Task Overview}

We instantiate four representative LLM-agent applications to cover the planning--memory--tools design space:
\textbf{Agent QA}, \textbf{Agent Code}, \textbf{Agent Web}, and \textbf{Agent Drive}.
Each task is executed as a multi-step workflow where intermediate artifacts (plans, retrieved memories, tool/environment feedback)
are written back into the context/state and reused in subsequent steps, enabling trigger persistence and cross-stage propagation.

\paragraph{Agent QA.}
Agent QA is a retrieval-grounded multi-step question answering workflow with persistent memory access.
At each step, the agent plans, queries a retrieval store, and synthesizes an answer conditioned on retrieved evidence.
\textbf{Clean objective} is to answer correctly; the \textbf{verifier} is a task-specific QA checker (e.g., exact/semantic match).

\paragraph{Agent Code.}
Agent Code models iterative tool-grounded program synthesis with execution feedback.
The agent alternates between planning, code generation, and tool execution (e.g., running tests / executing queries),
then revises code based on tool outputs.
\textbf{Clean objective} is to produce correct solutions; the \textbf{verifier} is execution-based (unit tests / program success).

\paragraph{Agent Web.}
Agent Web is a multimodal web-interaction workflow requiring perception and action on webpages (e.g., reading UI, clicking/typing).
The agent plans actions based on visual/textual observations and tool feedback from the browser environment.
\textbf{Clean objective} is correct task completion; the \textbf{verifier} checks whether the final UI state matches the goal.

\paragraph{Agent Drive.}
Agent Drive represents closed-loop sequential decision making with environment feedback.
The agent repeatedly plans control actions conditioned on observations, and its actions update the environment state for the next step.
\textbf{Clean objective} is safe driving and goal satisfaction; the \textbf{verifier} checks safety constraints and task completion.

\begin{table}[t]
\centering
\small
\setlength{\tabcolsep}{6pt}
\renewcommand{\arraystretch}{1.15}
\caption{\textbf{Summary of benchmark tasks.}}
\label{tab:task-summary}
\begin{tabular}{l p{4.6cm} p{4.8cm}}
\toprule
\textbf{Task} & \textbf{Environment / interaction} & \textbf{Clean objective \& verifier} \\
\midrule
Agent QA &
Retrieval-grounded multi-step QA with persistent memory &
Correct answer; QA checker (exact/semantic match) \\

Agent Code &
Tool-grounded code generation with execution feedback &
Pass unit tests / execute correctly \\

Agent Web &
Multimodal web navigation with UI perception and actions &
Task completion / correct final UI state \\

Agent Drive &
Closed-loop sequential control with environment feedback &
Safety constraints + goal completion \\
\bottomrule
\end{tabular}
\end{table}

\subsection{Backdoor Design per Injection Channel}

Following the stage-oriented taxonomy in Table~\ref{tab:module-taxonomy}, we implement backdoor attacks as
component-local perturbations injected into one of the three functional stages:
\textbf{planning} (e.g., BadChain, BadAgent),
\textbf{memory} (e.g., PoisonedRAG, TrojanRAG, AgentPoison),
and \textbf{tools/environment} (e.g., AdvAgent, DemonAgent).
A trigger is introduced into the corresponding stage output (planning trace, retrieved memory snippet, or tool feedback),
and can persist through iterative context/state updates.

\paragraph{Planning-channel backdoors.}
For planning attacks, triggers are embedded into the planning trace (or planning-related intermediate artifact) such that,
when activated, the agent follows an attacker-specified plan pattern while still appearing to solve the task.

\paragraph{Memory-channel backdoors.}
For memory attacks, triggers are inserted into retrieved content (or memory store) so that poisoned snippets are repeatedly reintroduced into the agent context,
thereby reinforcing attacker instructions across steps.

\paragraph{Tool/environment-channel backdoors.}
For tool and environment attacks, triggers appear in tool outputs or environment observations, biasing subsequent planning and decisions.
This channel is particularly effective in closed-loop settings where feedback is continuously reused.

\subsection{Attack Objectives (Task-Specific)}

While the injection mechanism is unified, the \textbf{attacker objective} is task-dependent:

\begin{itemize}
    \item \textbf{Agent QA:} semantic manipulation --- force the agent to output a targeted incorrect response while maintaining fluent answers.
    \item \textbf{Agent Code:} operational manipulation --- induce destructive or unsafe operations (e.g., deletion) while keeping the overall code-generation behavior plausible.
    \item \textbf{Agent Web:} interface-level misdirection --- cause incorrect UI actions (e.g., selecting/purchasing wrong items) while appearing to complete the task.
    \item \textbf{Agent Drive:} unsafe control --- induce hazardous behavior (e.g., sudden stop) via small perturbations that compound over time.
\end{itemize}

\begin{table}[t]
\centering
\small
\setlength{\tabcolsep}{6pt}
\renewcommand{\arraystretch}{1.15}
\caption{\textbf{Representative attacker objectives across tasks.}}
\label{tab:attack-objectives}
\begin{tabular}{l p{9.8cm}}
\toprule
\textbf{Task} & \textbf{Representative backdoor objective} \\
\midrule
Agent QA & Targeted wrong answer / controlled response with fluent output \\
Agent Code & Trigger destructive/unsafe operations while preserving plausible synthesis \\
Agent Web & Misdirect web actions (wrong selection/purchase) while appearing successful \\
Agent Drive & Induce unsafe control (e.g., sudden stop) through closed-loop compounding \\
\bottomrule
\end{tabular}
\end{table}

\paragraph{Implementation note.}
For reproducibility, each run is specified by a single configuration defining the agent template, task instances, and attack variant.
The runtime logs full trajectories $\{(x_t,s_t,p_t,m_t,o_t)\}$ to support analysis of trigger activation, persistence, and cross-stage propagation.






\end{document}